\documentclass{article}
\pdfoutput=1
\usepackage{arxiv}

\usepackage[utf8]{inputenc}
\usepackage{amsthm}
\usepackage{multirow}
\usepackage{subcaption}
\usepackage{mathtools}
\usepackage[export]{adjustbox}
\usepackage[T1]{fontenc}    % use 8-bit T1 fonts
\usepackage{url}            % simple URL typesetting
\usepackage{booktabs}       % professional-quality tables
\usepackage{amsfonts}       % blackboard math symbols
\usepackage{nicefrac}       % compact symbols for 1/2, etc.

\title{Navigating the Dynamics of Financial Embeddings over Time}
\author{Antonia Gogoglou, Brian Nguyen, Alan Salimov, Jonathan Rider, C. Bayan Bruss\\
Capital One,
McLean, VA 22102}
\date{June 2020}

\usepackage{amsmath}
\usepackage{multirow}
\begin{document}

\maketitle

%\renewcommand{\shortauthors}{Gogoglou, et al.}

%%
%% The abstract is a short summary of the work to be presented in the
%% article.
\begin{abstract}
    Financial transactions constitute connections between entities and through these connections a large scale heterogeneous weighted graph is formulated. In this labyrinth of interactions that are continuously updated, there exists a variety of similarity-based patterns that can provide insights into the dynamics of the financial system. With the current work, we propose the application of Graph Representation Learning in a scalable dynamic setting as a means of capturing these patterns in a meaningful and robust way. We proceed to perform a rigorous qualitative analysis of the latent trajectories to extract real world insights from the proposed representations and their evolution over time that is to our knowledge the first of its kind in the financial sector. Shifts in the latent space are associated with known economic events and in particular the impact of the recent Covid-19 pandemic to consumer patterns. Capturing such patterns indicates the value added to financial modeling through the incorporation of latent graph representations.
\end{abstract}

%%
%% The code below is generated by the tool at http://dl.acm.org/ccs.cfm.
%% Please copy and paste the code instead of the example below.
%%

%% A "teaser" image appears between the author and affiliation
%% information and the body of the document, and typically spans the
%% page.

%%
%% This command processes the author and affiliation and title
%% information and builds the first part of the formatted document.
%\maketitle

\section{Introduction}
\label{sec:intro}
Financial transactions, from credit card payments to stock purchases, can be viewed as edges on a graph where the nodes represent the parties involved in the transaction. This graph can contain features at various levels. For instance, the edges can contain information such as transaction amount and frequency while nodes themselves can contain rich features such as FICO score, income, and account balance. This structure lends itself to the application of Graph Neural Networks as shown previously in \cite{Bruss2019DeepTraxEG} and \cite{shumovskaia2020linking}.

At the same time, graphs of financial transactions have some unique properties not typical to the graphs used to develop GNNs. They can exhibit extreme power-law distributions. They are often heterogeneous, with a high level of variance in degree across nodes of different types. Additionally, they tend to be highly non-stationary in a multi-dimensional way. Firstly, new edges can form between nodes. Secondly, the weight of the edges can change over time. Finally, the set of nodes active on the network can grow and/or shrink. 

For the purpose of our work, we focus on credit card transactions. Credit card transactions form a bipartite graph between account holders and the merchants they shop at. Transactions are processed in a continuous stream, therefore to define this bipartite graph, one must look at a specific window in time defining $G_t$ for t amongst T discrete windows. This graph is defined over a set of accounts $V_ta$ \& merchants $V_tm$ active in the time window t and the weight placed on their edge $E_t$ can be defined as the frequency of transactions in that window. Between any two times $t-1$ and $t$ the following might be true:

\begin{itemize}
    \item $|V_t - V_{t-1}| > 0$: New accounts and merchants will appear 
    \item $|V_{t-1} - V_{t}| > 0$: Previously observed accounts and merchants will disappear
    \item $|E_t - E_{t-1}| > 0$: New edges will be observed
    \item $E_t \neq E_{t-1}$: Old edges will have their weight updated
\end{itemize}

Representation learning comprises a well-established set of techniques for embedding high cardinality sets (e.g. nodes in a graph or words in a vocabulary) in lower dimensional vector spaces in such a way that preserves their locality and co-occurence structure. Graph representation learning is often used as a generalized approach to feature generation from a graph structure that can be used in a number of down stream applications. In financial services these include fraud detection and credit decisions \cite{shumovskaia2020linking}, \cite{Bruss2019DeepTraxEG}. Many of the traditional representation learning techniques assume stationarity in the underlying structures that they embed. For instance, word embedding models are trained once on very large corpora and can be used for many downstream NLP tasks for years. Out-of-vocabulary words are often dropped or replaced with an unknown word token. Recent work has started to look at dynamic language and graph representation learning. However, the assumption of stationarity is only slightly relaxed as the datasets change slowly over time.

To solve the problem of graph representation learning on non-stationary financial graphs, this paper provides the following:
\begin{itemize}
    \item A description of our framework for training shallow embeddings from highly dynamic graphs over multiple timeframes that can empower down stream applications.
    \item An extensive in-depth qualitative analysis of embedding shift over time and identification of meaningful shifts.
    \item Association of temporal patterns in the representation space with real world transaction dynamics (e.g. shopping patterns, market changes) and merchant categories with a particular focus on the effects of Covid-19 pandemic to financial transactions.
    \item Time series analysis to filter rotational noise from the  dynamic retraining process and demonstration of how short-term
    representation shift can be effectively inferred from prior
    shift of the embedding space.
    \item In light of recent events related to the global pandemic, we investigate the ability of dynamic representations to quantify the extend of spending shifts that can prove useful in supporting decision makers in times of crisis.
\end{itemize}
Our analysis unveils the expressive power of embeddings for financial entities and their ability to encode the dynamics of consumer patterns in a dense representation fit for multiple downstream applications. 

\section{Related Works}
Graph representation learning seeks to embed the nodes of a network into a low dimensional vector space in such a way that the topological properties of the network are preserved \cite{perozzi2014deepwalk}, \cite{grover2016node2vec}, \cite{Kipf_gcn}, \cite{bronstein2017geometric}. Recent work investigating the application of these techniques to financial graphs shows promise as a general approach to modeling rich transaction networks \cite{Bruss2019DeepTraxEG} \cite{shumovskaia2020linking}. Much of the research in graph representation learning, including those applications in finance, have focused on static graphs. However, recent work into learning dynamic representations can be found in both the literature on word embeddings and as well as graph approaches.

\subsection{NLP Approaches}
Language is not static. New words are introduced, old words go out of fashion, and words can change their meaning. Research in NLP seeks to model the semantic changes in word usage over time. There are three main approaches explored within NLP for allowing words to evolve their meaning over time: chained initialization approaches, latent variable approaches and temporal alignment approaches. An important thing to note about dynamic word embeddings methods is that the time scales of change in language at a global level might be much longer and robust to exogenous influences compared to financial transaction graphs. 

Chained initialization approaches tackle the problem of embeddings over time pragmatically by modifying static word embedding models to allow them to update dynamically over time. The simplest approach is to use the embeddings of the previous time period as the intialization for the current time period and retrain on the current time period \cite{kim2014temporal} and \cite{stewart2017measuring}. Other variants use this same methodology but make use of word2vec's two embedding spaces, one for the observed words and one for the context, by fixing the context space over time and allowing the observed words to train each time period  \cite{Carlo2019TrainingTW}. The baseline methodology presented here for financial transactions most closely resembles these approaches.

Latent variable approaches take a Bayesian formulation of the problem assuming that the words observed in a corpus are draws from a latent distribution parameterized by the word embeddings. The Bayesian formulation allows them to explicitly model the dynamics of word evolution in a Markovian manner. In \cite{rudolph2018dynamic} vectors of words and their contexts are used to paramaterize a Bernoulli distribution which defines the probability of the word in that context. Similar to \cite{Carlo2019TrainingTW} the context vectors are fixed on some reference point, such as $t_0$, but the word vectors are allowed to update over time. Another approach proposed by \cite{bamler2017dynamic} is to treat the evolution of words as an Ornstein-Uhlenbeck process to ensure the evolutionary process doesn't stray too far from the origin. This work proposes two Kalman inference methods one for a streaming settings (filtering) and one where the entirety of the corpus is available (smoothing). We present our work utilizing Kalman filtering over the dynamic embeddings as a meta-model following chained initialization training. 

Both of these approaches seek to chain the training of individual word embeddings over time. A third approach taken by \cite{hamilton2016diachronic}, seeks to achieve alignment at a more global scale. In this approach embeddings at different time periods are trained independently of one another to allow them to capture the specific semantic structures within that time period. However, most embedding algorithms are not rotation invariant and therefore cannot be compared to one another when trained independently in this way. So this approach learns an alignment matrix between each time period to ensure the coordinate systems are aligned.

\subsection{Graph Approaches}

Graph approaches to dynamic graph problems have focused on solving the problem of transmitting changing graph information through time while also allowing full expressiveness at each time step. Tracking this temporal information by learning dynamic representation through time using recurrent architectures \cite{pareja2019evolvegcn, goyal2018dyngraph2vec} is one approach. In \cite{goyal2018dyngem} a combination of chained initialization and dynamically widening and/or deepening techniques on deep auto-encoders are employed to preserve stability across time steps. Additionally, the importance of community changes over time is underlined as a metric that captures the meaningful shift between snapshots of the graph. 
Other approaches seek to make explicit the temporal dependencies between graphs by injecting this information into the embeddings themselves. In \cite{kumar2019predicting} the concept of node trajectory is introduced and the embeddings of each snapshot are treated as instances of a  time-series of the overall trajectory of the node. Temporal dependencies are seen as a physical process and a context vector can be created to track this evolution and provide usable insight into shifting embeddings. \cite{trivedi2018dyrep} further explores the temporal dependencies between graphs by investigating long-term topological changes in graph structure and short term interactions between nodes, treating these short term interactions as a temporal point process model parameterized by an attention mechanism.

All of these approaches extend static graph approaches to dynamic ones by examining the relationships between $T$ graphs stacked on top of each other; by introducing novel architecture or changing existing architectures to maintain a temporal element they capture temporal dependencies effectively. However, the increase in the volume of information quickly makes large temporal graph problems intractable as more and more time steps are considered. Simplifications, such as restricting the change in size to the number of nodes, are necessary to make up for the increase in graph size as well as increase in parameters for more complicated models. 

\section{Methodology}
\label{sec-method}

\subsection{Generating Dynamic Embeddings}
As discussed in Section \ref{sec:intro}, there are particular challenges with the representation of highly dynamic graphs such as transaction graphs. To address these challenges we devise a training framework that ensures both coherence and flexibility throughout multiple time frames. The static baseline method for each snapshot of the transaction graph is described in  \cite{Bruss2019DeepTraxEG}. The incoming data are represented in tabular format where each data entry corresponds to a particular transaction between an account and a merchant associated with a given timestamp. A bipartite graph of accounts and merchants connected through transactions can subsequently be inferred from this dataset. As described in \cite{metapath2vec}, random walks can be generated across different types of nodes for heterogeneous graphs. Building off this concept and to enhance flexibility and training capacity we consider each node of a particular type as a {\it bridge} that facilitates a connection between two nodes of another type by interacting with both of them. For the case of merchant type nodes this results in {\it \{merchant,account,merchant\}} triplets that represent a connection between two merchants if the same account has interacted with both of them. Similarly, accounts are linked by performing transactions at the same merchant within the time window. This results into two homogeneous projections of the original heterogeneous graph. In the present work, we focus on connections between {\it Brand Level Merchants}, meaning that merchants are represented by their brand name without distinguishing between different locations of the same merchant. This approach results in a highly interconnected graph representing spending patterns across the nation. By tuning the time window appropriately this connectivity can be efficiently tuned.

This link construction approach can be viewed as formulating random walks of length two with a context window of one. The resulting pairs are subsequently fed into a skip-gram model \cite{Mikolov2013} to generate dense low dimensional representations for each node. We posit that given the noisiness and high interconnectivity in transactional data contemplating longer range interactions through larger walk lengths would constitute it challenging to guarantee that a negatively sampled merchant will not appear as part of the positive context as well. Considering a larger context window would pose the risk of interrelating merchants that are not actually meaningfully similar. The frequency of appearance for each training pair represents the strength of each link, i.e. the edge weight. 

In order to capture the state of the transaction graph at different moments in time, we form monthly snapshots at the end of each month following the aforementioned methodology. One issue that has been contemplated in literature \cite{Levy2014,Hamilton2017} is the random rotation of the embedding space in the skip-gram model. This rotation would constitute each snapshot's embeddings radically different in values compared to the previous and therefore their use in downstream models would be compromised. To address this matter, we opt for a warm-start training where each snapshot's model is initialized with the previous month's final state. This leads to nodes being cumulatively added to the embedding space. Once a merchant has entered the embedding space its position is maintained and receives updates only when new training pairs appear that include this merchant. Newly introduced nodes are added in the embedding space with a random initialization. 

{\bf Notation} Considering a set of $T$ snapshots which correspond to time stamps $t_0 < t_1<...<t_T$. For each time step $t_i$, $n_i^+$ positive context pairs are generated based on co-occurrence of transactions within a given time window, while $n_i^-$ pairs are sampled for negative context. In terms of probabilities the objective is to maximize the log likelihood of the positive context pairs being observed as opposed to the negative pairs:

\begin{equation}
    log(p^\pm | V_i) = \sum_{a,b}^{N}(n_{ab}^+ log(\sigma(v_a^T v_b)) + n_{ab}^- log(\sigma(-v_a^T v_b)))
\end{equation}

\begin{figure}[]
  \centering
  \scalebox{0.90}{
  \includegraphics[width=\textwidth]{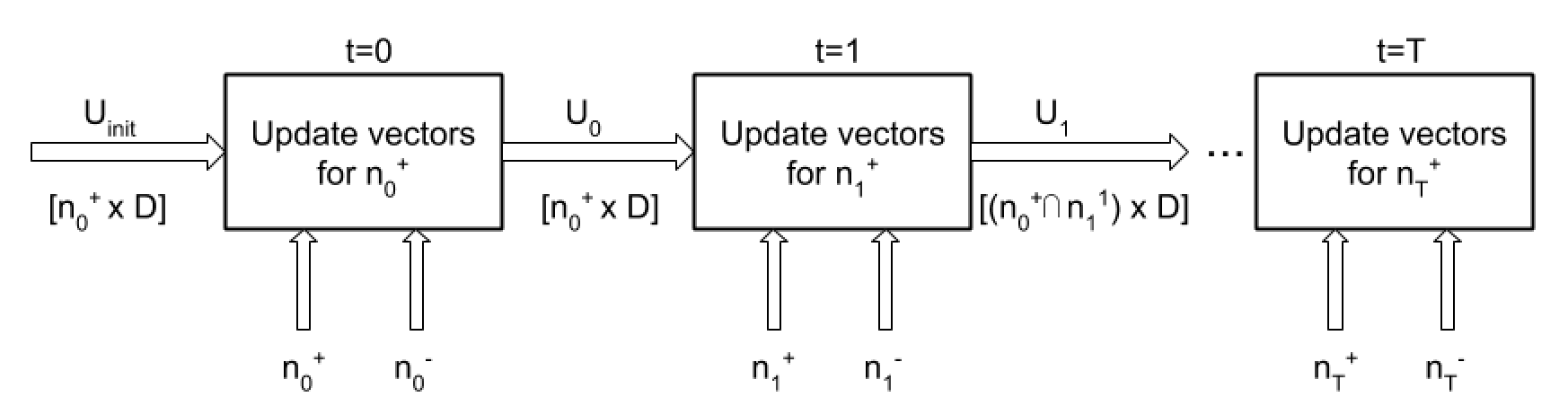}}
  \caption{Framework overview for dynamic training of skip-gram model for merchant embeddings with chained initialization over T sequential time stamps.} 
  \label{fig-workflow}

\end{figure}

Embedding lookup table $V_i \in R^{N \times d}$, where N is the cardinality of set $V_i$, is updated only for the nodes that are present in the $n_i^+$ context pairs of this month. However the number of data points $N$ in $V_i$ includes all the unique nodes that reside in the intersection of the sets of context pairs $[n_0^+,n_1^+,...,n_i^+]$ up to time stamp $i$. Details of training process are depicted in Figure \ref{fig-workflow}. 

{\bf Dataset} Our dataset consists of a set of {\it Brand Level Merchants} and their transactions over monthly snapshots between 2017-11 and 2020-03. For each merchant, we have an externally provided category code that identifies the type of business \footnote{https://www.iso.org/standard/33365.html}. Based on the dynamic training described above, $n_i^+$ averages at 50,000 per time stamp, while the total size $U_i$ of the embedding space extends to 260,000 merchants. 

\subsection{Quantification of Temporal Shifts}
The main focus of this work is to investigate the effectiveness of our dynamic training framework and explore the behavior of the resulting representations. In this section we will describe the methodology we designed to extract that knowledge from multiple snapshots of the embedding space including a series of downstream models that we applied on the sequence of representations.

{\bf Measuring shift} Since nodes in the latent space are represented as dense vectors, we select a set of first-order difference statistics to calculate their shift over time. The goal is to associate {\it representation shift} as measured in the latent space with real-world shift, i.e. {\it semantic shift}. Our selected metrics are {\it Euclidean distance} and {\it cosine distance} to measure both magnitude shift by the absolute values of the embedding dimensions but also similarity shifts as measured by the angle between embedding vectors. For a time stamp $t$ with $t \in [0,T]$, $\Delta_{magnitude}$ and $\Delta_{cosine}$ are defined as:

\begin{equation}
    \Delta_{magnitude} = ||v_t - v_{t-1}||_2
\end{equation}

\begin{equation}
    \Delta_{cosine} = 1 - \frac{v_t \cdot v_{t-1}}{||v_t||\cdot||v_{t-1}||}
\end{equation}

We proceed to investigate how these statistics change over time and what seasonal patterns arise from observing deviations from the average shift. Of particular interest is the timing of maximum shift for each merchant or a subset of them. For each merchant across a set of time stamps $T$, the maximum shift is defined as:

\begin{equation}
    max \Delta_{magnitude}(t) = argmax_t \Big(\frac{\Delta_{magnitude}}{\sum_t^{N} \Delta_{magnitude}}\Big)
\label{eq:maxMagn}
\end{equation}
The normalization by the cumulative shift of all the nodes $N$ within time stamp $t$ ensures that the shift ranking is sensitive to the nodes that drifted more compared to the drift of the rest of the nodes.

Another important measure of shift is the formulation of a node's neighborhood in the representation space. Over different time stamps and given the fact that nodes from a previous snapshot maintain their position in the space if they do not receive any updates, we expect the neighborhood to evolve but not deviate largely between consecutive snapshots. By defining neighborhood as the $k$ highest ranking in cosine similarity nodes around a target node, neighborhood shift can be quantified as:

\begin{equation}
    \Delta_{neighborhood}(k,\delta t) = \sum^N \Big( \frac{top{\it k}_t \cap top{\it k}_{t-\delta t}}{top{\it k}_{t-\delta t}}\Big) / N_i
\label{eq:neigh}
\end{equation}

where $\delta t$ is the number of time stamps elapsed between two different calculations of the top-k neighborhoods for every node and varying values of $k$. Per node shifts are subsequently aggregated across all nodes $N$.

{\bf Shift trajectories} In order to formalize the quantification of shift over time we aggregate the first order statistics across different timeframes that differ by $\delta t$ and create a time series of shifts:
\begin{equation}    
    \tau_{\Delta_{cosine}} = \{ \Delta_{cosine}(t_0),\Delta_{cosine}(t_0+\delta_t),...,\Delta_{cosine}(T)\}
\end{equation}

The resulting time series may be viewed as a {\it state space} model where individual components such as a financial trend component or a seasonal one are combined to produce the observed values of shift. By decomposing the time series into its constituent components we aim to identify the effects of the trend component as well as the expected random component that arises from the rotation of the embedding space after multiple rounds of Skip-gram training.

We employ filtering of the time series which entails the estimation of current values of the state from past and current observations. A particular method of time series filtering that has been utilized in conjunction with embedding approaches is {\it Kalman filtering} \cite{kumar2019predicting, bamler2017dynamic}.
The Kalman filter operates on a series of measurements observed over time, e.g. the embedding shift, and assumes that they contain some level of Gaussian noise or inaccuracy, using the time series to estimate $P(x_t | z_{0:T-1})$ as given by the equations below. 
\begin{equation}
    x_{t+1} = A_xt + b_t + \mathcal{N}(0, Q_t)
\label{eq:kalman_x}
\end{equation}
\begin{equation}
   z_t = C_xt + d_t + \mathcal{N}(0, R_t)
\label{eq:kalman_z}
\end{equation}

%x̂ n,n= x̂ n,n−1+Kn(zn−x̂ n,n−1)=(1−Kn)x̂ n,n−1+Knzn
%% https://pykalman.github.io/#pykalman.KalmanFilter cite
Building on that assumption, it produces estimates that tend to be more accurate than those based on individual measurements alone. Therefore it can provide an estimation of the true shift trajectory across time stamps. To explore trajectories themselves, we compute neighborhood $n$ based on Equation \ref{eq:neigh}. We applied Kalman smoothing to embeddings (normalized using $\Delta_{magnitude}$)
elementwise, by assuming an independent difference vector, namely {\it velocity}, for each embedding component and calculating the next time step components as a linear combination of predicted and actual values calculated by an EM algorithm. We therefore compute the normalized difference vector $\Delta\hat{v_t} \in \mathcal{R}^d$ or {\it velocity}, where $d$ is the dimension of the embedding vectors. 
\begin{equation}
 \Delta\hat{v_t} = \hat{v}_t - \hat{v}_{t-1}
\label{eq:velocity_v}
\end{equation} 

{\bf Predicting shift} Finally, a predictive sequence model is applied to attempt inference of future shifts given past observations. If there are meaningful short-term and/or long-term dependencies then a sequence model should be able to identify them and outperform a naive average based baseline. Additionally, by altering the size of the considered sequence we discover the shift rate of transaction patterns and the duration of trends. A Long Short Term Memory (LSTM) model for regression \cite{Oancea2013} is utilized to predict shift in $\Delta_t$ time stamps in the future by looking back at $l$ time stamps ({\it sequence length}) and over a training period of $t_{tr}$ slices which represents the {\it training length} on which the model is trained:

\begin{equation}
     \Delta_{cosine}(t) =
     f_{tr}([\Delta_{cosine}(t-l),\Delta_{cosine}(t-l+1),...,\Delta_{cosine}(t-1)])
\end{equation}
 
\section{Results}
Utilizing the metrics and methods described in Section \ref{sec-method} we perform a rigorous qualitative analysis of the temporal evolution of merchant behavior as represented in the embedding space. The objective of this analysis is threefold. First, if seasonal patterns and well known financial events are encoded in the embedding space, that increases the applicability and effectiveness of the proposed representation. Second, we aim to quantify meaningful shift that represents changes in transaction patterns as opposed to the arbitrary embedding space rotation and ensure the consistency of the representation over time. Finally, by treating the sequence of the aforementioned statistics over different snapshots as a time series we seek to address the predictability of this shift.
% For the purposed of our experiments, merchants that were not part of more than $50\%$ of the time slices considered in this analysis were removed from the statistics calculation

\subsection{Shift distribution}
\begin{figure}[]
  \centering
  \scalebox{0.9}{
  \includegraphics[width=\textwidth]{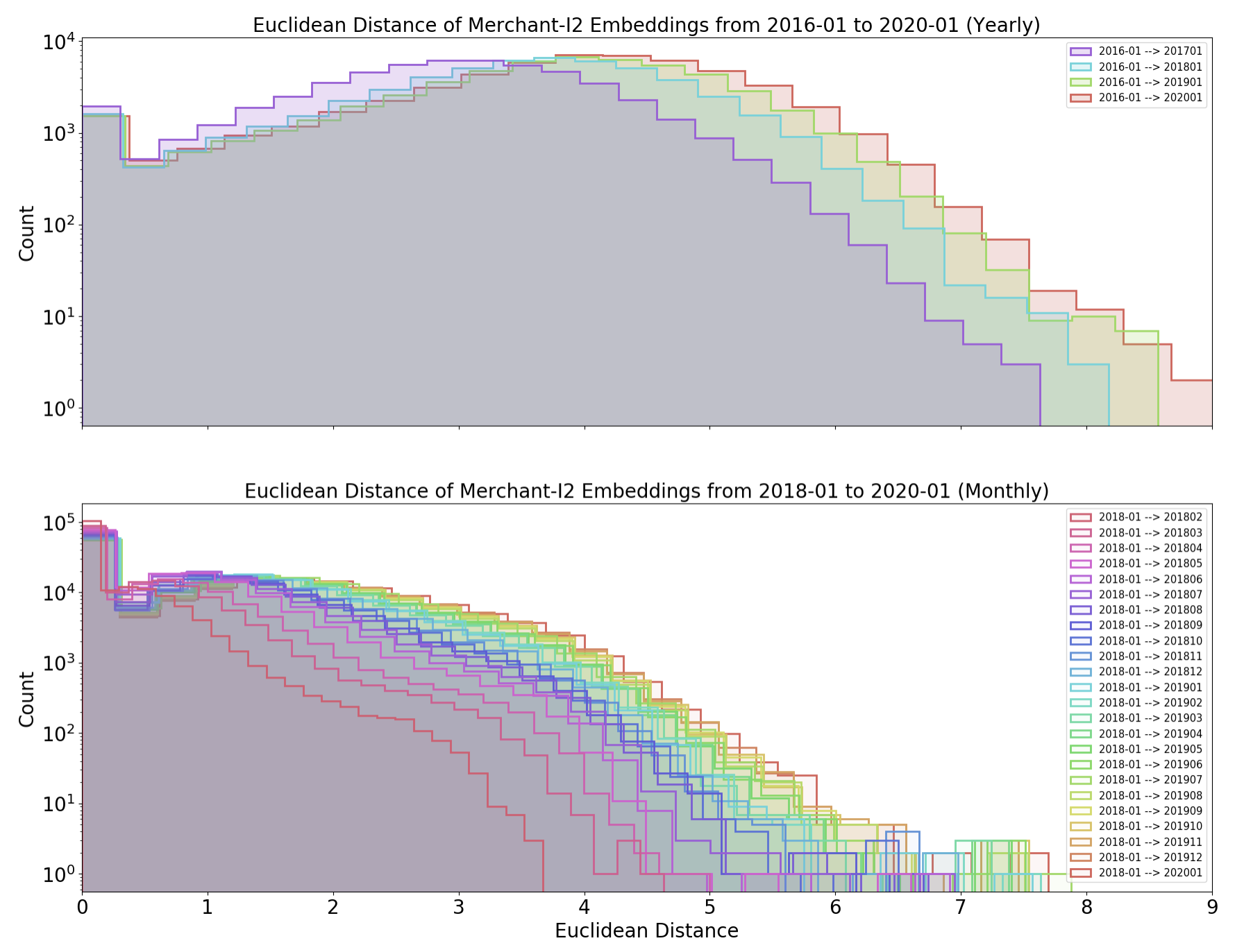}}
  \caption{Distribution of $\Delta_{magnitude}$ over different time frames on a yearly and monthly cadence indicating smooth temporal shifts}
  \label{fig-distHist}
\end{figure}

\begin{figure}[]
  \centering
  \scalebox{0.9}{
  \includegraphics[width=\textwidth]{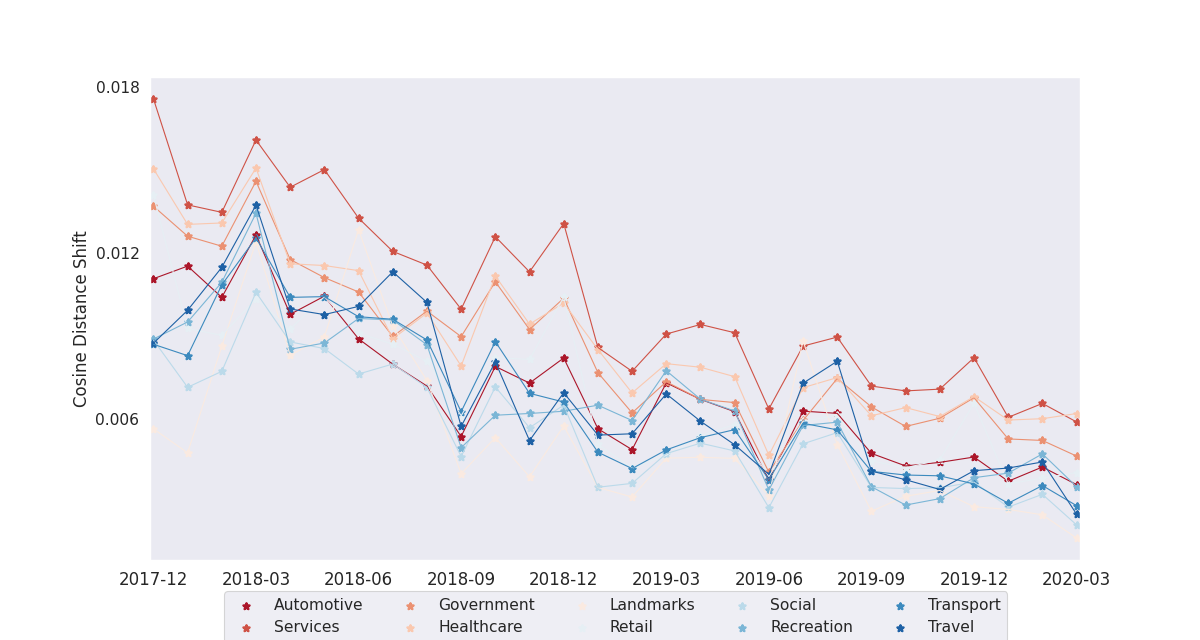}}
  \caption{Average $\Delta_{cosine}$ over all time stamps for different merchant categories indicating how the size of shifts differs significantly across types of merchants}
  \label{fig-cosineShift}
\end{figure}

{\bf Representation shift} Figures \ref{fig-distHist} and \ref{fig-cosineShift} visualize the distribution of shifts across different time frames. Based on Figure \ref{fig-distHist} that depicts $\Delta_{magnitude}$ on yearly and monthly cadence, we can see that embedding vectors gradually move away from their original position as time progresses indicating the smoothness of the embedding trajectories and a directed motion. We notice that $\Delta_{cosine}$ also reduces over time (Figure \ref{fig-cosineShift}) however the distribution is still far from stationary. As more nodes are added and maintain their position in the embedding space, the amount of shift in more recent time frames is mostly attributed to change of transaction patterns while in the original time frames shift is observed due to the first time appearances of merchants. Additionally, we observe that even though the general trends are shared across categories of merchants, meaning spikes and valleys usually happen across all categories at the same time frames, there are some categories that steadily demonstrate higher deltas. For instance Government, Services and Travel appear to maintain higher $\Delta_{cosine}$ across all time frames, probably due to seasonal volatility experienced in Travel and Government while Services represent a high percentage of the merchant space thus incorporate the changes for a wide range of nodes.

\begin{figure}[]
  \centering
  \scalebox{0.9}{
  \includegraphics[width=\textwidth]{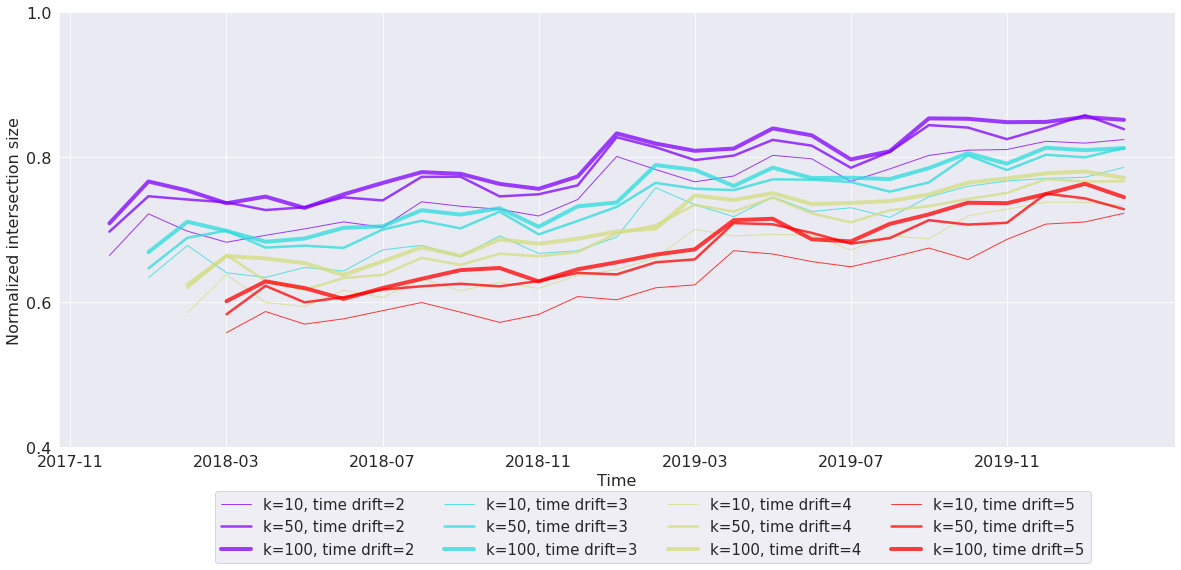}}
  \caption{Normalized intersection size between {\it topk} neighborhoods over each time stamp and past ones with varying time drift.}
  \label{fig-intersect}
\end{figure}

Since the latent representations are trained to bring closer together the most similar nodes, it is interesting to see how the ranking of the closest neighbors evolves. Therefore we measure $\Delta_{neighborhood}(k,\delta_t)$ from Equation \ref{eq:neigh} for k=10,50 and 100 and $\delta_t$=2,3 and 4. Results are exhibited in Figure \ref{fig-intersect}. As expected the longer the time that elapsed, the smaller the intersection between the different versions of a merchant's neighborhood. However, the shift seems to be mostly in the closest of neighbors (k=10) whereas the broader neighborhood (k=100) remains roughly stable. This can be attributed to general shopping patterns being relatively similar over the years with small seasonal trends that alter the ranking within the neighborhood. In Table \ref{tab:neighs} we observe specific examples of the nodes that belong in the intersection of neighborhoods from four different time stamps for two well known merchants, namely the Ritz-Carlton hotel and the Banana Republic retailer. As can be observed the intersection between 2019-10 and 2019-12 is generally higher for both compared to a longer time window of 5 months (2019-10 to 2020-03) indicating that similarity shifts happen gradually over time.

\begin{figure}[]
  \centering
  \scalebox{0.9}{
  \includegraphics[width=\linewidth]{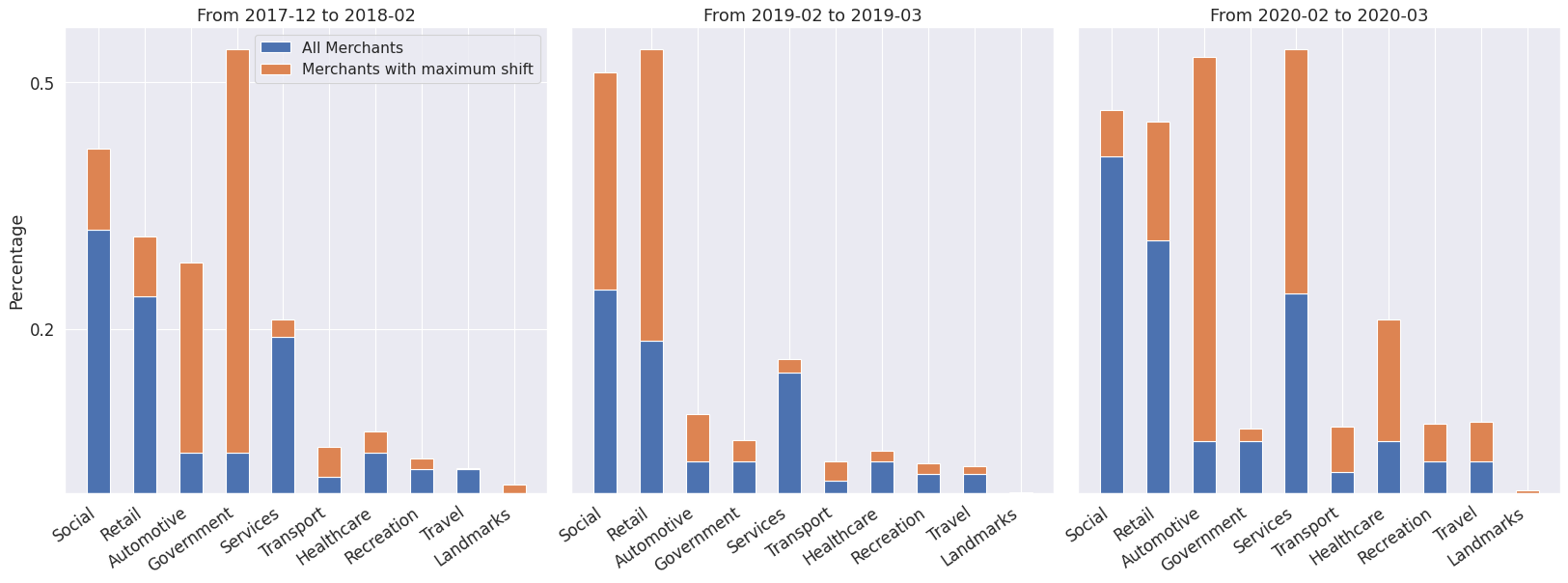}}
  \caption{Percentages of different merchant categories in the set of most frequent merchants and in the set of top 10,000 shifting merchants showing the periods when a category changes more than the rest}
  \label{fig-percCat}
\end{figure}

{\bf Semantic Shift} The question arises whether all merchant embeddings shift at the same pace or whether there exist categories that showcase larger shifts in particular time periods. We measure the percentage of each merchant category and repeat this calculation amongst the top 10,000 shifting merchants (according to $\Delta_{cosine}$) between different time stamps. In Figure \ref{fig-percCat} we report results for time periods of interest. For this experiment, the set of merchants for each time stamp has been trimmed based on frequency of transactions. Therefore only nodes that received significant updates in this training round are included. The first chart showcases the effect of seasonal spending patterns such as holiday seasons that cause the categories of Retail and Social to exhibit the largest shift between December and February. Services as well as Travel follow a similar trend. An interesting pattern appears in the next two charts, by comparing the shift in 2019-03 and 2020-03. It appears that Social, Retail and Services achieve higher percentages amongst the maximum shifting merchants compared to the same time last year, which is in line with the effects of the Covid-19 pandemic in consumer behavior. 

Similar patterns appear when we explore the maximum shift in magnitude $max\Delta_{magnitude}$ from Equation \ref{eq:maxMagn}. From the time series of $\Delta_{magnitude}$ for each merchant, the month in which the merchant exhibited the highest shift compared to the total magnitude shift of all merchants in that same month is selected as {\it max shift month}. By displaying the counts of merchants that exhibit their max shift in each given month, we notice trends appearing that coincide with known financial events. In the beginning of 2019 and then again in the summer of 2019 market volatility may have influenced consumer behavior. Interestingly, counts appear to spike in the first few months of 2020 with a peak in March 2020 when the Covid-19 pandemic caused major changes in spending patterns.

\begin{table}[]
\caption{Examples of {\it topk} neighbors for time drift equal to 2 (top row) and 5 months (bottom row)}
\label{tab:neighs}
\centering
%\resizebox{\textwidth}{!}{
\scalebox{0.99}{
\begin{tabular}{ccll}
\hline
Timeframes                          & \multicolumn{1}{l}{k size} & The Ritz-Carlton                                                                                                                                                           & Banana Republic                                                                                                                                                               \\ \hline
\multirow{2}{*}{2019-10 to 2019-12} & 5                          & \begin{tabular}[c]{@{}l@{}}Link Restaurant Group\\ La Meridien\end{tabular}                                                                                                & \begin{tabular}[c]{@{}l@{}}Apple Store\\ Ugg\end{tabular}                                                                                                                     \\ \cline{2-4} 
                                    & 50                         & \begin{tabular}[c]{@{}l@{}}Four Seasons Hotel\\ Nopsi Hotel\\ Double Tree\\ Ruby Slipper Cafe\\ Link Restaurant Group\\ Miami Beach Resort\\ Superior Seafood\end{tabular} & \begin{tabular}[c]{@{}l@{}}Saks Fifth Avenue \\ Swarovski\\ Kiehl's Since 1851\\ Apple Store\\ Lululemon\\ Abercrombie and Fitch\\ David's tea\\ Nordstrom\\ Ugg\end{tabular} \\ \hline
\multirow{2}{*}{2019-10 to 2020-03} & 5                          & \multicolumn{1}{c}{Link Restaurant Group}                                                                                                                                  & Gap                                                                                                                                                                           \\ \cline{2-4} 
                                    & 50                         & \begin{tabular}[c]{@{}l@{}}Four Seasons Hotel\\ Nopsi Hotel\\ The Daily Beet\\ Link Restaurant Group\end{tabular}                                                          & \begin{tabular}[c]{@{}l@{}}Gap,\\ Neiman Marcus\\ Kiehl's Since 1851\\ Nike Outlet\\ Bath \& Body Works\end{tabular}                                                          \\ \hline
\end{tabular}}
\end{table}

\begin{figure}[]
  \centering
  \scalebox{0.9}{
  \includegraphics[width=\linewidth]{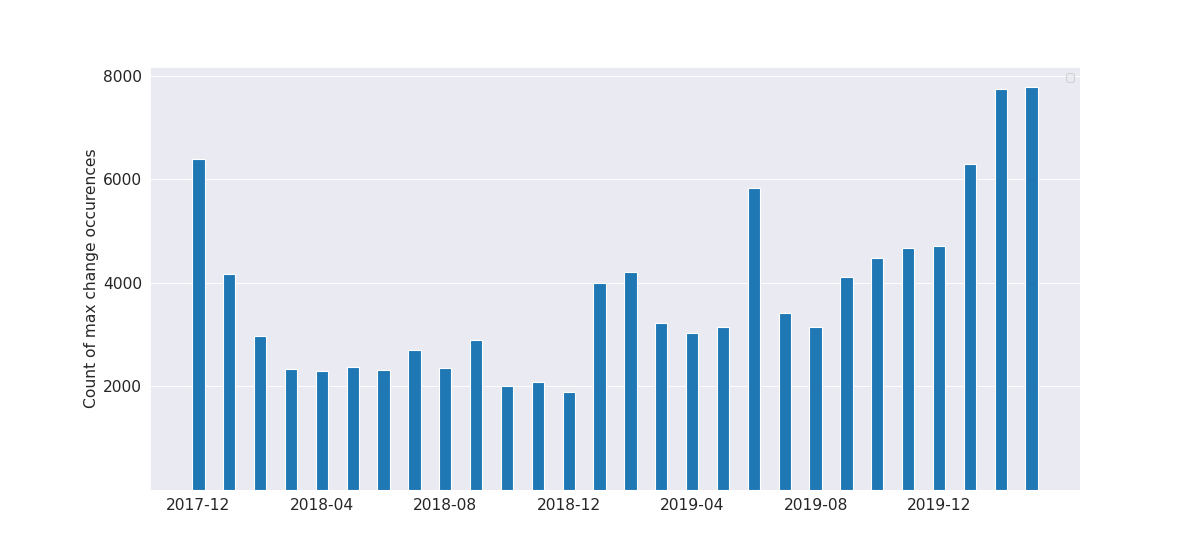}}
  \caption{Normalized count of merchants that exhibit their maximum shift in each time stamp.}
  \label{fig-maxChange}
\end{figure}

\begin{figure}[]
  \centering
  \includegraphics[width=\textwidth]{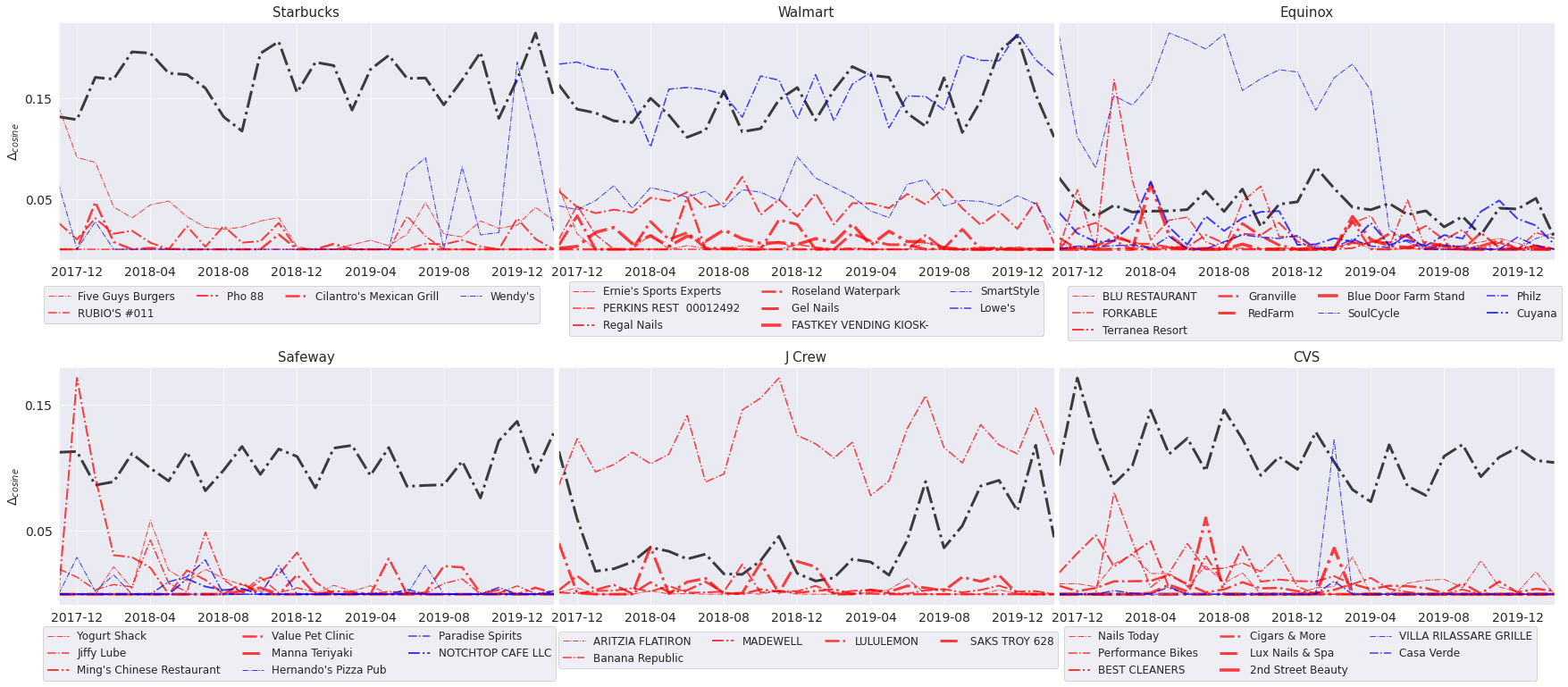}
  \caption{Cosine shift of seed node (depicted in black) and the respective shifts of their neighbors from the first time stamp (red) and the last one (blue) comparing the trajectories of seed node and neighbors}
  \label{fig-cosNeigh}
\end{figure}

Moving towards a microscopic view of the relationship between representation shift and neighborhood evolution we explore individual merchants and how their {\it topk} neighbors across different time stamps relate to their $\Delta_{cosine}$. Results are depicted in Figure \ref{fig-cosNeigh}. We consider the neighbors in the first time stamp (2017-11) and the last one (2020-03). If a node belongs to the original neighbor set then it is depicted in red even if they belong to the final set of neighbors as well. The neighboring nodes represented with blue lines however are selected so that they are not part of the original set of neighbors. We search for {\it topk} neighbors with k=10, but we remove the ones that exhibited unchanged $\Delta_{cosine}$ for more than 70\% of the considered time stamps. Two patterns arise from this analysis: Firstly, seed nodes tend to have generally high similarity in representation shift with their neighbors (e.g. J Crew with Banana Republic or Walmart with Lowe's). Secondly, merchants that emerge later as neighbors of the seed node tend to be more volatile and have started following similar shift trends with the seed during intermediate time stamps. Equinox and Starbucks, for instance, with their neighbors SoulCycle and Wendy's appear to observe common trends during the last 5 time stamps. These findings suggest that $\Delta_{cosine}$ correlates with meaningful changes in transaction patterns that alter neighborhood formulation. 

\subsection{Shift Trajectories as Time Series}
{\bf Visualizing and filtering trajectories} Moving away from selecting the biggest shifts from the $\Delta_{cosine}$ time series, we also explore the sequence as a whole and attempt to identify its constituent components.
To explicitly visualize patterns in both magnitude and direction of embedding shift, we calculate 120 different 2 hop neighborhoods with $k=100$ neighboring nodes. After normalizing these embeddings using $\Delta_{magnitude}$, we smooth them using Kalman smoothing on a per-merchant basis and take their element-wise difference per time stamp to generate velocity vectors. 

\begin{figure}[]
 \centering
 \scalebox{0.9}{
 \includegraphics[width=\textwidth]{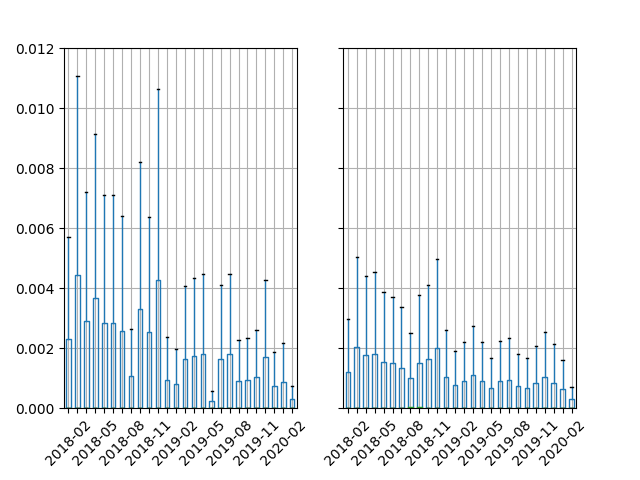}}
 \caption{Box plot of average cosine differences between merchant embeddings in each time step without smoothing on the left and with Kalman smoothing on the right.}
 \label{fig-raw-kalman-velocity-cosine}
\end{figure}

We first note that according to Figure \ref{fig-raw-kalman-velocity-cosine}, smoothing did lower both the mean and variance of cosine distance across time stamps. We believe this indicates that arbitrary embedding shift was lowered operating on the assumption that embeddings follow a linear trajectory in embedding space, as assumed in the Kalman filter. This follows the results from \cite{kumar2019predicting, bamler2017dynamic} where a similar concept was employed during embedding calculation rather than afterwards. We benefit from filtering after embedding generation rather than beforehand because the filter can be agnostic to the source of the data. It also allows us to use both embeddings calculated using well-established embedding algorithms on massive data set with more downstream flexibility pertinent on the use case.

\begin{figure}[]
 \centering
 \includegraphics[width=\linewidth]{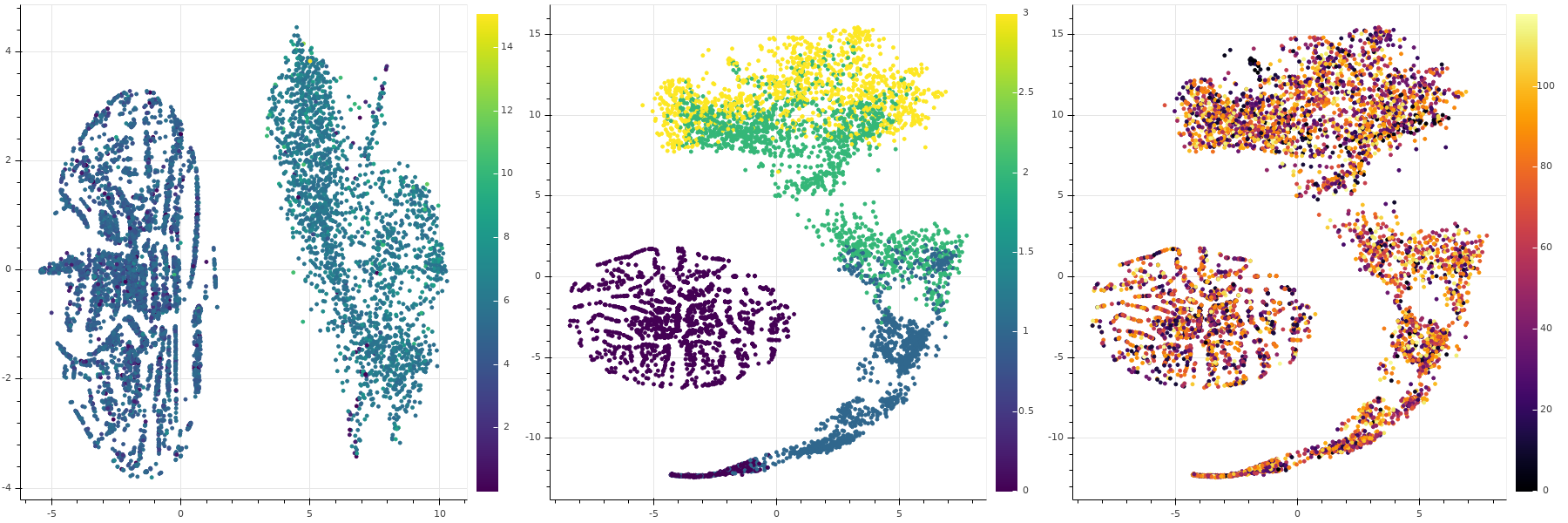}
 \caption{t-SNE of 2020-03 velocity vectors shaded by log frequency, quartile cosine distance, and neighborhood membership out of 120 neighborhoods. Center and right are Kalman smoothed, while left chart is raw embeddings.}
 \label{fig-kalman-velocity-viz}
  \end{figure}

Subsequently, the velocity vectors $\Delta\hat{v_t}$, from equation \ref{eq:velocity_v}, of all merchants across all time stamps are aggregated and passed through t-SNE dimensionality reduction. Dynamic t-SNE is a challenging problem and its limitations have been discussed in literature \cite{boystov2017dynamictnse}. In this work we opt to calculate t-SNE coordinates for all time stamps at once so that comparisons are enabled across different snapshots of the embedding space. As we observe in Figure \ref{fig-kalman-velocity-viz}, for the non smoothed velocity embeddings two clusters emerge that correspond to regions of high and low frequency. After smoothing (second chart), the frequency pattern was maintained and by shading according to $\Delta_{cosine}$ quartiles we notice higher degree of separation among the clusters high frequency region. In these clusters, the denser ones appear to be correlated with high $\Delta_{cosine}$. Finally, in the last chart of Figure \ref{fig-kalman-velocity-viz} we observe that neighborhood memberships expand across regions of zero and non-zero movement (low and high frequency) which indicates that the inclusion of non-updating nodes in our sequential training workflow allows nodes to maintain connections from previous time frames and expand their neighborhood in both highly changing and more stable regions. Both small and large merchant neighborhoods have this tenancy; when reduced to a single merchant neighborhood, a large retailer or a single parking meter both show momentum from previous snapshots even if their neighbors are speeding away from their original positions in the high cosine distance area. 
%  Moreover using the highly scalable implementation \cite{chan2019gpu} we were able to perform the calculations for all 200,000 nodes in $<3 $ minutes, using $perplexity=500$. Perplexity values were selected according to the power law to consider global structure rather than local ones using \cite{chan2019gpu}.
 
% \begin{figure*}[]
%   \centering
%   \includegraphics[width=\textwidth]{cosine_neighs2.png}
%   \caption{Cosine shift of seed node (depicted in black) and the respective shifts of their neighbors from the first time stamp (red) and the last one (blue)}
%   \label{fig-cosNeigh}
% \end{figure*}

{\bf Predicting trajectories} Finally, we explore the predictability of representation shift as measured by $\Delta_{cosine}$ and attempt to identify the effect of long- and short-term dependencies in the latent space. As discussed in Section \ref{sec-method}, we employ an LSTM model trained with multiple past time stamps and a sequence of previous $\Delta_{cosine}$ values available at each time stamp. Since the transaction space is highly dynamic, extending the sequence length beyond a certain point could be adding little value to the estimation of the next shift value. Therefore sequence length for our experiments varies between 1 and 7 months, meaning $\Delta_{cosine}$ of each month with the previous one are available for 1 to 7 months. Training length lies in the same range to investigate if a large number of training points is needed to make effective predictions. A naive baseline is added as a frame of reference, where we estimate the next month's $\Delta_{cosine}$ as a moving average of the past sequence values. For performance evaluation Mean Square Error is reported. 
Results are shown in Tables \ref{tab-201912} and \ref{tab-202002} for the test month of 2 different time periods. Steady increase in performance is observed over the baseline for all sequence and training lengths which indicates that non trivial temporal and perhaps repetitive patterns arise in the embedding space. It appears that extending the sequence length up to the 5 previous time steps achieves the highest performance; however increasing the sequence length offers limited to no benefit. This could be attributed to seasonal trends that move embeddings further than their representation 7 time steps ago as well as the overall rotation of the embedding space. Moreover, error values are on average higher for March 2020 when an unprecedented change in consumer behavior occurred due to the Covid-19 pandemic.

\begin{table}[h]
\caption{Mean Square Error ($\times 10^4$) for naive baseline model and LSTM with different timeframes}
\label{tab-201912}
\begin{minipage}{.5\linewidth}
\centering
\caption{Training up to 2019-12 and test 2020-01}
\begin{tabular}{llllll}
\hline
\multicolumn{1}{c}{\multirow{3}{*}{\begin{tabular}[c]{@{}c@{}}sequence \\ length\end{tabular}}} & \multicolumn{5}{c}{training length}                     \\ \cline{2-6} 
\multicolumn{1}{c}{}                                                                            & \multirow{2}{*}{Baseline} & \multicolumn{4}{c}{LSTM}  \\ \cline{3-6} 
\multicolumn{1}{c}{}                                                                            &                           & 1    & 3    & 5    & 7    \\ \hline
1                                                                                               & 1.75                      & 1.08 & 0.95 & 0.94 & 0.92 \\ \hline
3                                                                                               & 1.52                      & 0.68 & 0.64 & 0.62 & 0.61 \\ \hline
5                                                                                               & 1.37                      & 0.62 & 0.56 & 0.56 & {\bf 0.47} \\ \hline
7                                                                                               & 1.38                     & 0.63 & 0.59 & 0.58 & 0.50 \\ \hline

\end{tabular}
\end{minipage}
%\end{table}
%\begin{table}[h]
%\caption{Mean Square Error ($\times 10^4$) for naive baseline model and LSTM with training month up to 2020-02 and test month 2020-03}
%\label{tab-202002}
\begin{minipage}{.5\linewidth}
\centering
\caption{Training up to 2020-02 and test 2020-03}
\begin{tabular}{llllll}
\hline
\multicolumn{1}{c}{\multirow{3}{*}{\begin{tabular}[c]{@{}c@{}}sequence \\ length\end{tabular}}} & \multicolumn{5}{c}{training length}                     \\ \cline{2-6} 
\multicolumn{1}{c}{}                                                                            & \multirow{2}{*}{Baseline} & \multicolumn{4}{c}{LSTM}  \\ \cline{3-6} 
\multicolumn{1}{c}{}                                                                            &                           & 1    & 3    & 5    & 7    \\ \hline
1                                                                                               & 1.38                      & 0.95 & 0.91 & 0.91 & 0.89 \\ \hline
3                                                                                               & 1.30                      & 0.70 & 0.69 & 0.70 & 0.69 \\ \hline
5                                                                                               & 1.21                      & 0.69 & 0.68 & 0.68 & {\bf 0.62} \\ \hline
7                                                                                               & 1.26                      & 0.70 & 0.69 & 0.68 & 0.63 \\ \hline

\end{tabular}
\end{minipage}
\end{table}

\section{Conclusion}
This work introduces a dynamic scalable graph representation workflow deployed on financial graphs of transactions with accounts and merchants as entities. The resulting representations across a range of timestamps are evaluated for consistency and the concept of semantic shift is introduced as a means to measure the effectiveness of latent representations in real world application scenarios. We perform a multi-step qualitative analysis that is to our knowledge the first of its kind in the financial sector to extract trends and patterns from the evolution of the merchant representations over time.

\bibliographystyle{unsrt}
\bibliography{main}

\end{document}